%% file: main.tex
\documentclass[runningheads]{llncs}

\usepackage{eccvabbrv}
\usepackage{graphicx}
\usepackage{booktabs}
\usepackage[accsupp]{axessibility}
\usepackage{hyperref}
\usepackage{orcidlink}
\usepackage[section]{placeins}
\usepackage{caption}
\usepackage{graphicx}
\usepackage{booktabs}
\usepackage{amssymb}
\usepackage{algorithm}
\usepackage{algorithmic}
\usepackage{float}

\title{GeoRoute: Geometry-Aware Hybrid Inference for Traffic Future-Frame Prediction}

\titlerunning{GeoRoute}

\author{
Khang Minh Le\inst{1} \and
Hieu Dinh Trung Pham\inst{1} \and
Danh Thanh Luu\inst{2, 5} \and 
Tien Nam Le\inst{3, 5} \and
Hieu Anh Ngo\inst{1} \and 
Phuong Huu Vu Tran\inst{1} \and 
Son Nguyen Minh Le\inst{1} \and 
Nghia Trong Nguyen\inst{4,5} \and 
Tu Tran Thi Cam \inst{4,5} \and
Huy Minh Nhat Nguyen\inst{1} \and 
Cuong Tuan Nguyen\inst{1}
}

\authorrunning{K.~M.~Le et al.}

\institute{
PAMI Lab, Vietnamese-German University, Vietnam 
\and
University of Science, Ho Chi Minh City, Vietnam
\and
Ho Chi Minh City University of Technology, Vietnam \and
University of Information Technology, Ho Chi Minh City, Vietnam \and 
GenAI4E Lab
}

\begin{document}

\maketitle

\input{sec/0_abstract}

\input{sec/1_intro}
\input{sec/2_related_work}
\input{sec/3_method}

\input{sec/4_experiment}

\input{sec/6_conclusion}

\bibliographystyle{splncs04}
\bibliography{ref}

\end{document}

%% file: sec/0_abstract.tex
\begin{abstract}
Long-horizon future-frame prediction is important for autonomous driving,
traffic surveillance, and intelligent transportation systems, yet remains
challenging due to temporal ghosting, geometry drift, and inconsistent object
motion. Recent latent video diffusion models have achieved impressive visual
quality, but directly applying them to structured traffic scenes often leads to
unstable geometry and degraded temporal coherence over extended horizons. We
present a training-free inference framework that stabilizes reliable static
structure in pretrained video predictions through multi-frame temporal context
and view conditioned routing. For front-camera videos, our method refines generated futures with a
multi-frame depth-based point-splatting renderer that projects static content from observed
history frames while preserving dynamic regions from the generative base model.
For heterogeneous traffic views, a frozen vision-language model infers a coarse
camera group from the observed clip and selects a specialized motion-based
predictor. The framework requires neither retraining nor fine-tuning of the
underlying video model and can be applied directly to pretrained generators. We
validate the proposed framework on the AI City Challenge Track~5 benchmark \cite{aicity2026track5},
where our final system ranks fifth on the official full-test leaderboard. The
leaderboard results show competitive overall low-level fidelity, while
qualitative comparisons support the intended static-structure effect. The
framework does not change the original generator architecture. Our code will be available on \href{https://github.com/Khangle-2006/AICITY_Track_5}{Github}.

\keywords{
Future-Frame Prediction,
Traffic Video Forecasting,
Video Diffusion Models,
Training-Free Inference,
Geometry-Aware Refinement,
Temporal Consistency
}
\end{abstract}

%% file: sec/1_intro.tex
\section{Introduction}
\label{sec:introduction}

Future-frame prediction is important for intelligent transportation systems,
autonomous driving, and traffic surveillance. Given a short observed video clip,
the goal is to predict future frames that preserve scene geometry, object
motion, and temporal coherence. This problem has been studied through
recurrent, predictive-coding, motion-aware, and flow-based video prediction
methods~\cite{srivastava2015unsupervised,lotter2016prednet,predrnn2017,villegas2017hierarchical,liu2017deepvoxel,liang2017dualmotiongan,denton2018svg,gao2022simvp}. In traffic
scenarios, the task is especially challenging due to ego-motion, independently
moving vehicles and pedestrians, occlusions, and diverse camera
viewpoints~\cite{yu2020bdd100k,wts2024}.

Recent latent video diffusion models have greatly improved visual generation
quality by extending latent diffusion with temporal layers, spatiotemporal
attention, or transformer backbones~\cite{rombach2022ldm,ho2022videodiffusion,voleti2022mcvd,blattmann2023videoldm,blattmann2023svd,videocrafter,animatediff,hacohen2024ltxvideo,kong2024hunyuanvideo}. However, directly applying
pretrained video generators to long-horizon traffic forecasting often produces
temporal ghosting, afterimages, geometry drift, and inconsistent object motion.
Training-free video generation and editing methods also face similar
frame-to-frame consistency challenges~\cite{khachatryan2023text2videozero,yang2023rerender,geyer2023tokenflow,fatezero2023}.

Fine-tuning a large video generator on the target benchmark is expensive,
data-sensitive, and difficult under challenge constraints. Training-free
diffusion control methods adapt pretrained models by manipulating attention,
propagating features, or enforcing cross-frame consistency~\cite{prompt2prompt,masactrl,pix2video,khachatryan2023text2videozero,geyer2023tokenflow}, but
they do not directly address the static-geometry artifacts that appear in
traffic future-frame prediction. Geometry and motion cues offer complementary
signals: optical flow can propagate visual content~\cite{liu2017deepvoxel,raft},
while monocular depth and semantic segmentation help separate static scene
structure from dynamic actors~\cite{ranftl2020midas,ranftl2021dpt,chen2017deeplabv3}.

We propose a training-free inference framework for long-horizon traffic
future-frame prediction. For front-camera driving videos, we use a pretrained
video generator as an appearance and dynamic-content prior, then refine static
regions with a multi-frame depth-based point-splatting renderer. Depth and
actor masks are estimated from observed history frames, while
history-to-generated-view alignment is obtained by matching static features
between an observation and each generated base frame. Reliable static pixels
are then projected using z-buffer splatting. Dynamic
or low-confidence regions are preserved from the generative base prediction.
Thus, the geometry branch stabilizes static structure rather than re-estimating
future actor trajectories.
For other visual regimes, we use specialized motion-based predictors, since
elevated or nearly fixed views often favor deterministic propagation over
generic front-camera generation.

Our framework operates entirely at inference time and requires neither
retraining nor fine-tuning of the underlying video generator. We evaluate on AI
City Challenge Track~5, using BDD front-camera
videos~\cite{yu2020bdd100k} and WTS traffic views~\cite{wts2024}. The benchmark
reports PSNR and SSIM~\cite{wang2004ssim} for reconstruction fidelity,
LPIPS~\cite{zhang2018lpips} for perceptual similarity, CLIP
similarity~\cite{radford2021clip} for semantic alignment, and
FID/FVD~\cite{heusel2017fid,unterthiner2018fvd} for image and video
distribution quality. Our final system achieves a challenge score of
\textbf{73.28}.

Our contributions are summarized as follows:
\begin{itemize}
    \item We propose a training-free inference framework for long-horizon
    traffic future-frame prediction that improves static-geometry stability and
    low-level structural fidelity without modifying model parameters or sampling
    schedules.
    \item We introduce a confidence-aware geometry refinement module with multi-frame
    history support. The module projects reliable static pixels from observed frames
    into future views, while preserving dynamic and low-confidence regions from the
    generative base prediction.
    \item We design a Qwen-assisted view-conditioned routing strategy that combines
    front-camera generative refinement with view-specific motion-based
    predictors.
    \item We validate the framework on AI City Challenge Track~5 \cite{aicity2026track5}, achieving a
    final challenge score of \textbf{73.28}.
\end{itemize}

%% file: sec/2_related_work.tex
\section{Related Work}
\label{sec:related_work}

\subsection{Latent Video Diffusion Models}
Diffusion models have become a dominant paradigm for visual generation, while
latent diffusion reduces the cost of pixel-space denoising by operating in a
compressed representation~\cite{rombach2022ldm}. Video Diffusion Models and
MCVD extended diffusion to temporal generation and conditional future-frame
prediction~\cite{ho2022videodiffusion,voleti2022mcvd}. Subsequent latent video
models introduced temporal U-Net modules, latent-space processing, and
image--video joint training to improve resolution and efficiency
~\cite{blattmann2023videoldm,videocrafter,blattmann2023svd,animatediff}.
More recent systems such as HunyuanVideo and LTX-Video employ transformer-based
spatiotemporal architectures for scalable video synthesis and stronger
conditioning alignment~\cite{kong2024hunyuanvideo,hacohen2024ltxvideo}.
Despite these advances, inference-only application to long-horizon traffic
prediction can still produce temporal ghosting, geometry drift, and
inconsistent object motion.

\subsection{Training-Free Diffusion Control and Video Editing}
Training-free diffusion control reuses representations within pretrained
models without updating their parameters. Prompt-to-Prompt manipulates
cross-attention maps, while MasaCtrl shares mutual self-attention features to
preserve spatial layout and appearance~\cite{prompt2prompt,masactrl}. For
videos, Text2Video-Zero introduces cross-frame attention and latent motion
dynamics without video-model training~\cite{khachatryan2023text2videozero}.
Pix2Video, Rerender-A-Video, FateZero, TokenFlow, and VidToMe propagate
attention, diffusion features, correspondences, or tokens across frames to
improve editing consistency~\cite{pix2video,yang2023rerender,fatezero2023,geyer2023tokenflow,vidtome2024}.
These approaches primarily target text-guided generation or editing. GeoRoute
instead refines future predictions with external geometry recovered from the
observed history and leaves the pretrained sampling process unchanged.

\subsection{Temporal Consistency and Geometry-Guided Refinement}
Optical flow provides dense temporal correspondences and is widely used for
warping or propagating visual content. RAFT estimates flow through recurrent
all-pairs matching, while FlowVid combines flow guidance with spatial
conditions to reduce errors caused by imperfect correspondence
~\cite{raft,flowvid2024}. Nevertheless, flow becomes unreliable under
occlusion, disocclusion, rapid camera motion, and independently moving actors.

Geometry-based view synthesis provides a complementary way to preserve scene
structure. SynSin projects predicted 3D features into novel views, layered
depth representations explicitly model occluded scene content, and splatting
operators resolve multiple source samples that map to the same target location
~\cite{synsin2020,shih2020threedphoto,niklaus2020splatting}. Inspired by these
principles, GeoRoute projects only masked static pixels from multiple observed
frames, resolves visibility with a z-buffer, and uses confidence-aware blending
to preserve generated content where geometry is unreliable.

\subsection{Hybrid Prediction for Heterogeneous Traffic Views}
Traffic datasets contain substantially different camera configurations.
BDD100K primarily provides front-facing driving videos with ego-motion, whereas
WTS includes vehicle, fixed, and overhead viewpoints
~\cite{yu2020bdd100k,wts2024}. Classical optical flow and adaptive background
models remain useful when camera geometry is stable
~\cite{farneback2003,zivkovic2004}. A single predictor may therefore be
suboptimal across all views. GeoRoute uses Qwen-assisted view-conditioned routing:
front-camera videos receive generative prediction followed by geometric
refinement, while traffic-camera views use conservative motion propagation.

%% file: sec/3_method.tex
\section{Method}
\label{sec:method}

\subsection{Problem Formulation}
\label{sec:problem_formulation}

Let $x_t$ denote the RGB frame observed at time $t$. Given a history of $T$
frames, $\{x_1,\ldots,x_T\}$, our goal is to predict the next $K$ faicity2026track5
$\{\hat{y}_1,\ldots,\hat{y}_K\}$. Thus, $x_T$ is the last observation and
$\hat{y}_j$ is the prediction at future step $j$.

Traffic prediction requires two different capabilities. Static structures such
as roads, lane markings, curbs, and buildings should remain geometrically
stable, whereas vehicles, pedestrians, and newly visible regions require
plausible synthesis. These requirements become harder under ego-motion,
occlusion, and heterogeneous camera viewpoints. GeoRoute therefore separates
the problem into two responsibilities: a pretrained generator proposes future
appearance and dynamic content, while an explicit geometry module corrects only
static regions for which observed evidence is reliable. Views with nearly
stable camera geometry use a simpler motion-propagation branch instead.

\subsection{Framework Overview}
\label{sec:method_overview}

Figure~\ref{fig:pipeline} shows the complete data flow. GeoRoute receives an
observed clip and its available textual description.
A frozen Qwen2.5-VL \cite{bai2025qwen25vl} model first classifies the observed clip into a coarse view
group and selects the corresponding prediction branch. Front-camera driving
videos pass through prompt construction, LTX-Video generation \cite{hacohen2024ltxvideo}, and
geometry-aware refinement. Traffic-camera videos pass through one of two
motion-based predictors according to the predicted view group.

\begin{center}
\includegraphics[width=\linewidth]{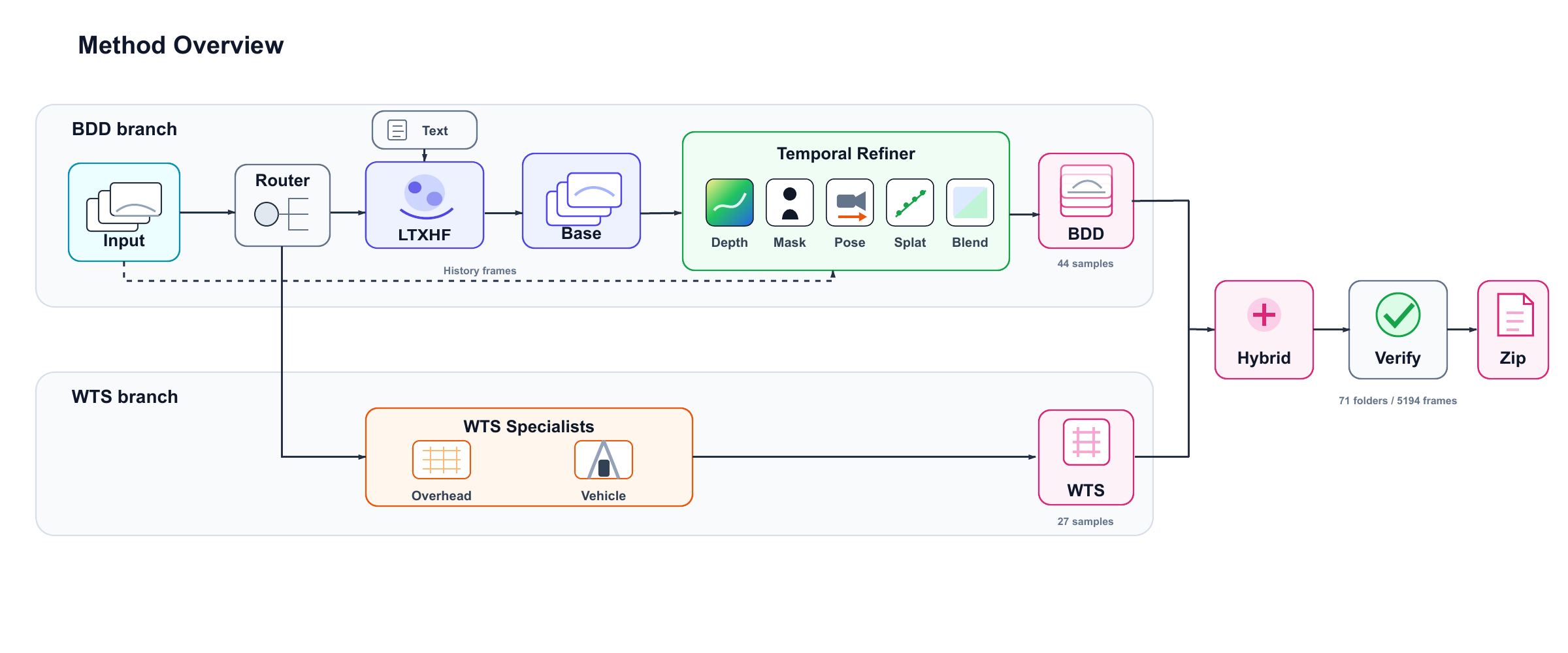}
\captionof{figure}{\textbf{GeoRoute overview.}
Qwen infers a coarse view group from the observed clip and routes it to either
the text-conditioned LTX-Video \cite{hacohen2024ltxvideo} and geometry-refinement branch or a stable-view
motion predictor. Each branch returns the same sequence of future frames.}
\label{fig:pipeline}
\end{center}

This design follows the physical behavior of the input rather than assuming
that one predictor is optimal for every camera. Front-camera videos contain
strong perspective change and disocclusion, so a generator is needed for
content that cannot be copied from the past. However, its output may contain
ghosted lane boundaries or drifting buildings. GeoRoute anchors those static
regions to real observations. For overhead or slowly moving traffic cameras,
the observed background is already a strong future reference; deterministic
propagation avoids unnecessary generative drift. All routing rules are fixed
before evaluation, and no future ground-truth frame is used.

\subsection{View-Conditioned Routing}
\label{sec:view_aware_inference}


The frozen Qwen2.5-VL-7B-Instruct \cite{bai2025qwen25vl} router independently matches the first,
middle, and last observed frames of each video to one of three operator-oriented
visual-regime prototypes: (i) forward-driving views, (ii) stable elevated or
fixed views, and (iii) other oblique roadside or mobile views. They select,
respectively, generative refinement, median-background actor propagation, or
conservative region-normalized flow propagation. Routing therefore depends on
visual behavior rather than dataset identity or physical camera protocol.

Following the camera patterns represented by BDD100K and
WTS~\cite{yu2020bdd100k,wts2024}, the instruction uses observable cues:
viewpoint and road geometry, cross-frame background displacement, and whether
motion is global or actor-local. A low forward-facing viewpoint with a road
vanishing point and scene-wide background displacement indicates front-camera
driving. An elevated viewpoint with a nearly stationary background and
actor-local motion indicates an overhead/fixed camera; remaining oblique
roadside or mobile views are assigned to the third regime. Qwen \cite{bai2025qwen25vl} receives no
dataset identity, metadata, file name, target frame, or manual label. The
prototypes, prompt, parser, and branch mapping are fixed for all samples, so
clips with similar viewpoint and motion structure receive the same route. This
is label-free zero-shot routing; no clustering model is fitted on the test set.
An unparsable response falls back to the conservative third branch.
Figure~\ref{fig:view_routing} shows representative unlabeled inputs.

\begin{center}
\includegraphics[width=0.92\linewidth]{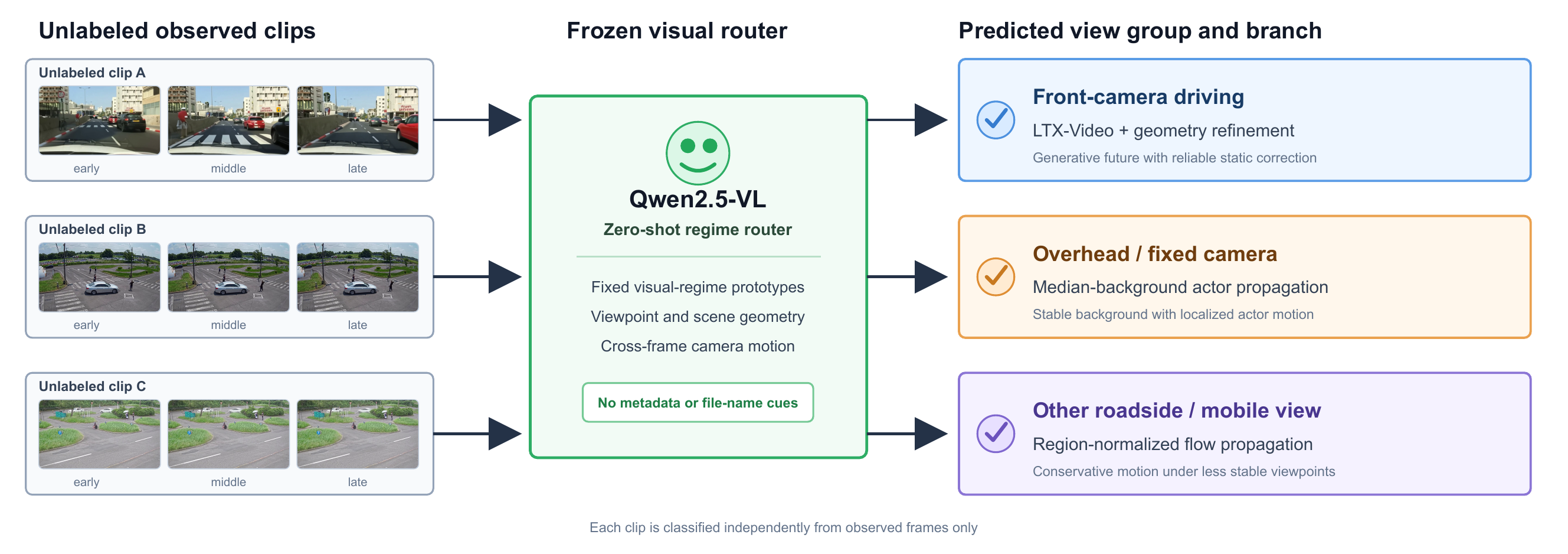}
\captionof{figure}{\textbf{Qwen-assisted view routing.}
Three observed frames are matched to fixed visual-regime prototypes; the
selected regime determines the corresponding inference branch.}
\label{fig:view_routing}
\end{center}

\subsection{Qwen-Assisted Prompt and Base Prediction}
\label{sec:text_prompt_construction}

For a front-camera sample, the base generator is conditioned on both the
observed clip and a compact prompt. Frozen Qwen2.5-VL-7B-Instruct \cite{bai2025qwen25vl} summarizes
sparsely sampled history frames together with the supplied text and predicted
view group~\cite{bai2025qwen25vl}. Its structured output is restricted
to visible scene layout, actor categories, and coarse motion context; it does
not infer exact future trajectories. Sampling across the history reduces the
influence of transient occlusion.

Figure~\ref{fig:prompt_construction} illustrates this decomposition. The view,
scene, actor, and motion fields are inserted into a deterministic template. A
short negative field discourages common generation artifacts such as temporal
ghosting, duplicated actors, warped road markings, flicker, and abrupt viewpoint
changes. Empty fields in the fixed schema fall back to generic traffic-scene
phrases rather than invented details.

\begin{center}
\includegraphics[width=0.92\linewidth]{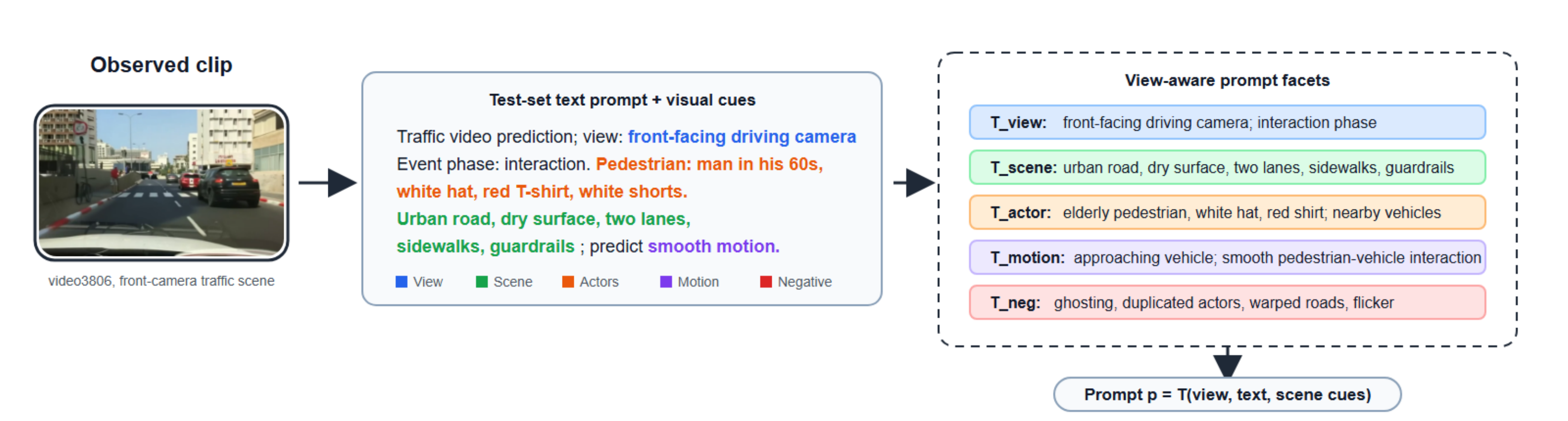}
\captionof{figure}{\textbf{Qwen-assisted prompt construction.}
Observed frames and the available sample text are summarized into grounded
view, scene, actor, and motion cues, then assembled by a fixed template.}
\label{fig:prompt_construction}
\end{center}

Qwen remains frozen and is used only at inference time. Its view output selects
the branch, while its scene facets condition pretrained
LTX-Video~\cite{hacohen2024ltxvideo} within the front-camera branch; geometry
refinement does not use language predictions. Let $b_j$ be the generated frame
at future step $j$. It provides the default full-frame prediction, including
dynamic and disoccluded content that cannot be projected from history. Later
modules replace only confident static regions and otherwise retain $b_j$.
GeoRoute therefore claims improved static-geometry stability and low-level
structural fidelity, not explicit recovery of actor trajectories.

\subsection{Geometry-Aware Static Refinement}
\label{sec:geometry_refinement}

Figure~\ref{fig:bdd_refiner} expands the refinement stage. Its purpose is
simple: recover reliable background pixels from real history frames, move them
to the predicted camera view, and leave all other pixels to the generator.

\begin{center}
\includegraphics[width=0.95\linewidth]{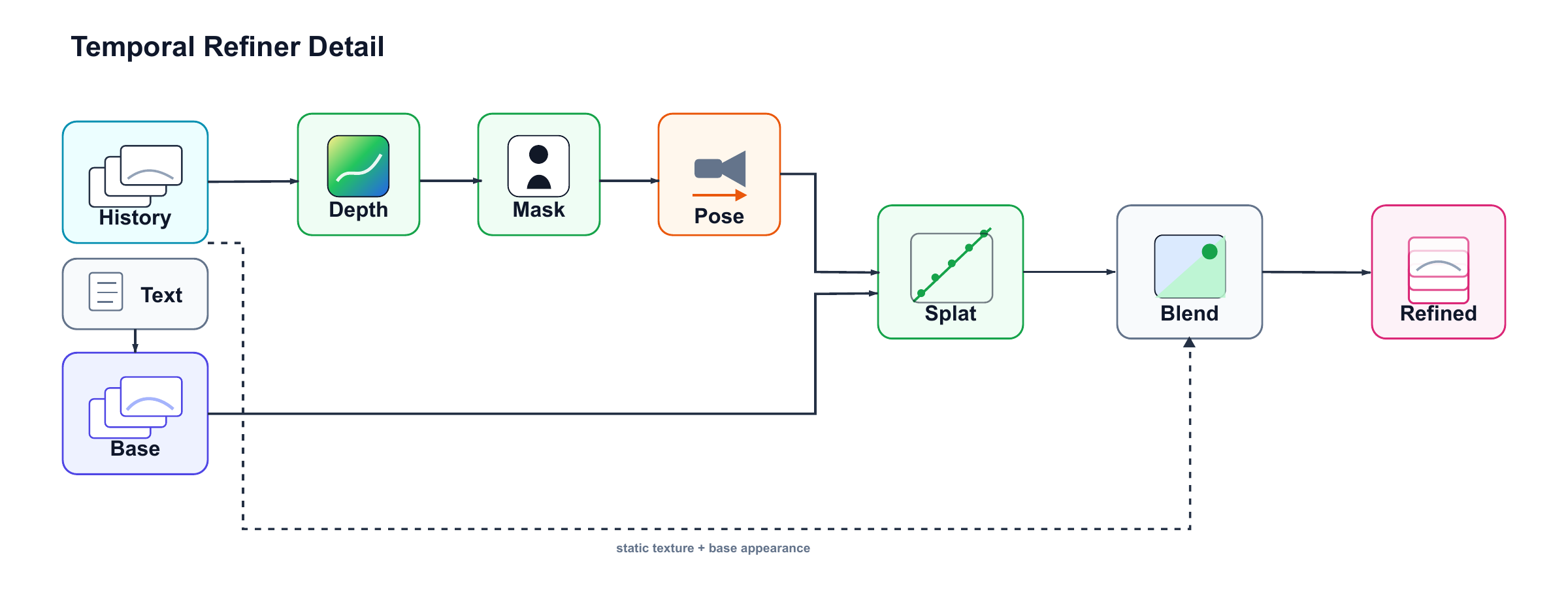}
\captionof{figure}{\textbf{Geometry-aware static refinement.}
Depth, actor masks, and camera motion turn observed history frames into a
rendered static layer. Confidence-aware blending applies this layer only where
the projection is reliable.}
\label{fig:bdd_refiner}
\end{center}

\paragraph{Selecting static observations.}
We select up to eight history frames, ordered from old to recent. The most
recent frame is usually the closest visual reference, while older frames may
contain background regions that later become disoccluded. For each selected
frame, DPT Hybrid-MiDaS estimates relative depth
~\cite{ranftl2020midas,ranftl2021dpt}, and DeepLabV3-ResNet50 estimates an actor
mask~\cite{chen2017deeplabv3}. Person, bicycle, motorcycle, car, bus, and related
vehicle classes are excluded from static projection. We also mask actors in
each base frame so that a projected road or building cannot overwrite a
generated vehicle.

The two masks have different roles: the history mask prevents past actors from
being copied to outdated positions, while the base-frame mask prevents projected
background from overwriting actors generated for the future.

Depth is used only as relative geometry. The predicted inverse depth is
normalized with robust image percentiles and converted to a bounded positive
range. Because calibrated intrinsics are unavailable, we use an approximate
pinhole camera whose principal point is the image center. These approximations
are sufficient for conservative local projection, but they are not interpreted
as metric 3D reconstruction.

\paragraph{Aligning a history frame to a future view.}
For every future base frame $b_j$, ORB features are matched between static
regions~\cite{rublee2011orb}. Matches on actors are discarded. The remaining
points, together with history depth, are used by EPnP-RANSAC
~\cite{lepetit2009epnp,fischler1981ransac} to estimate the camera transform. To
make the geometric operation explicit, a static pixel $u$ with depth $D(u)$ is
mapped to the future view as
\[
\tilde{u}' \sim K\!\left(R\bigl(D(u)K^{-1}\tilde{u}\bigr)+t\right).
\]
Here, $\tilde{u}$ and $\tilde{u}'$ are the pixel coordinates before and after
projection, $K$ is the approximate camera matrix, and $(R,t)$ is the estimated
relative pose. This equation is used only for static pixels. A pose is rejected
when there are fewer than 20 ORB matches or 12 RANSAC inliers. These rejected
raw estimates may be filled from the pose sequence before the method falls back
to $b_j$.

Specifically, interpolation concerns the transform from the most recent history
frame to each future base frame. When a raw pose is missing and at least two
future steps have valid poses, rotation is represented by its three-component
Rodrigues vector and concatenated with the three translation components. Each
of the six components is linearly interpolated over the future-step indices,
with the nearest valid pose extended at sequence boundaries, and the resulting
sequence is smoothed by a triangular window of length seven. The implementation
does not impose an additional maximum-gap threshold; instead, every filled pose
is down-weighted to $Q_j=0.35$. If fewer than two valid poses are available, no
interpolation is performed and a missing step retains the base frame $b_j$.

Older history frames are first aligned to the most recent history frame, then
their transforms are composed with the most-recent-to-future transform. This
puts all selected observations in one future coordinate system without fitting
an unrelated pose for every history--future pair.

\paragraph{Rendering and multi-frame fusion.}
Each aligned history frame produces a candidate background layer. When several
source pixels land at the same target location, $z$-buffer splatting keeps the
closest point~\cite{synsin2020,niklaus2020splatting}. Candidate layers are then
combined from old to recent. A recent observation replaces an older one where
both cover the same pixel; an older observation is used only to fill a location
that newer frames do not cover. The resulting rendered background for future
step $j$ is denoted by $r_j$. Pixels with no projected observation are copied
from $b_j$, so the renderer never creates an empty hole.

This recent-first rule is deliberately conservative. It avoids averaging
slightly misaligned projections and gives older frames one clear role: expanding
coverage in static regions that are absent from the latest observation.

\subsection{Confidence and Final Blending}
\label{sec:confidence_blending}

Projection alone is not reliable enough to replace the generated image. Depth
may be noisy, pose may be weak, and projected colors may disagree with the
future appearance. We therefore compute a confidence value $w_j(u)$ for every
pixel $u$. Its four factors are shown directly in the equation:
\[
w_j(u)=
\underbrace{C_j(u)}_{\text{covered}}
\underbrace{\bigl(1-A_j(u)\bigr)}_{\text{static}}
\underbrace{\exp\!\left(-\Delta_j(u)/32\right)}_{\text{color agreement}}
\underbrace{Q_j}_{\text{pose quality}}.
\]
$C_j(u)$ equals one when at least one history frame covers the pixel and zero
otherwise. $A_j(u)$ is the actor mask of the base frame. $\Delta_j(u)$ is the
mean absolute RGB difference between $r_j(u)$ and $b_j(u)$ on the 8-bit color
scale. Finally, $Q_j$ is 1 for an accepted pose, 0.35 for a pose obtained by
temporal interpolation, and 0 when no pose is available. Thus, a pixel receives
high confidence only when it is covered, static, photometrically compatible,
and supported by a usable camera transform.

Coverage rejects rendering holes, the actor term protects dynamic content,
color agreement suppresses misaligned projections, and pose quality rejects
unsupported camera transforms. Their product is conservative because one weak
factor is sufficient to reduce the refinement weight.

We erode and blur this map to remove isolated projection boundaries, then apply
an exponential moving average across future steps to reduce flicker. The final
actor mask is applied once more after smoothing to prevent confidence from
leaking onto moving objects.

The refined prediction is a convex blend of the generated base frame and the
rendered static layer:
\[
\hat{y}_j(u)=
\bigl(1-\alpha w_j(u)\bigr)b_j(u)
+\alpha w_j(u)r_j(u),
\qquad \alpha=0.75.
\]
This expression has a direct interpretation. If $w_j(u)=0$, the output is the
LTX-Video prediction \cite{hacohen2024ltxvideo}. If confidence increases, more observed static content is
used, up to the maximum strength $\alpha$. The method therefore stabilizes
roads, curbs, lane markings, and buildings without claiming to correct an
incorrect actor trajectory already generated in $b_j$.

The blend remains soft even at high confidence. Retaining part of $b_j$ reduces
small exposure and color differences between observed and generated content,
while gradual confidence boundaries avoid visible seams around lane markings,
curbs, and actor masks. This is more stable than hard replacement, which can
turn a small projection error into a sharp spatial artifact.

\subsection{Motion Prediction for Stable Views}
\label{sec:motion_based_prediction}

Traffic-camera views use observed image motion instead of LTX-Video generation \cite{hacohen2024ltxvideo}.
Both predictors estimate Farneback optical flow~\cite{farneback2003} between the
last two observed frames and extrapolate it with a decaying displacement. The
two variants differ in what they preserve.

For overhead and fixed cameras, a per-pixel temporal median provides a stable
background. Flow magnitude and deviation from this median identify compact
moving regions. These actor regions are propagated forward and composited over
the unchanged median background. Large connected regions are rejected because
they are more likely to represent illumination or global camera change than a
single actor.

For vehicle-mounted and IP-camera views, global motion is more common, so a
hard actor/background decomposition is less reliable. We instead convert flow
magnitude into a soft regional mask. The last observed frame is warped by an
accumulated, clipped, and decayed flow, then blended with the unwarped frame
through this mask. This preserves local motion while limiting long-horizon
distortion. Exact thresholds, decay factors, and displacement limits are given
in Sec.~\ref{sec:experimental_setup}.

Neither stable-view predictor learns object trajectories. They extrapolate
observed image-space motion under a conservative background prior and are used
only for camera patterns where that assumption is appropriate.

\subsection{Inference Procedure}
\label{sec:inference_procedure}

Algorithm~\ref{alg:hybrid_inference} summarizes the method without introducing
additional notation.

\begin{algorithm}[H]
\caption{Training-free GeoRoute inference}
\label{alg:hybrid_inference}
\begin{algorithmic}[1]
\REQUIRE Observed clip and available text
\ENSURE Predicted future frames
\STATE Classify the observed clip with frozen Qwen2.5-VL-7B-Instruct
\IF{the sample is a front-camera driving view}
    \STATE Build a grounded prompt with frozen Qwen2.5-VL-7B-Instruct
    \STATE Generate base future frames with LTX-Video
    \STATE Estimate history depth and actor masks
    \STATE Estimate the recent-history-to-future pose sequence
    \STATE Interpolate and smooth missing poses when at least two are valid
    \FOR{each future step $j=1,\ldots,K$}
        \IF{a direct or filled pose is available}
            \STATE Render and fuse the multi-frame static layer $r_j$
            \STATE Compute the projection confidence $w_j$
            \STATE Blend $r_j$ with the base frame $b_j$ to obtain $\hat{y}_j$
        \ELSE
            \STATE Keep the base prediction $\hat{y}_j=b_j$
        \ENDIF
    \ENDFOR
\ELSIF{the sample is an overhead or fixed traffic-camera view}
    \STATE Apply median-background actor-layer propagation
\ELSE
    \STATE Apply conservative region-normalized flow propagation
\ENDIF
\RETURN The predicted future sequence
\end{algorithmic}
\end{algorithm}

The complete pipeline is training-free after the pretrained components are
loaded. It uses only the observed clip and available text; no target future
frame is used during routing, prompting, rendering, or blending.

%% file: sec/4_experiment.tex
\section{Experiments}
\label{sec:experiments}

\subsection{Experimental Setup}
\label{sec:experimental_setup}

We evaluate GeoRoute on the AI City Challenge Track~5 benchmark \cite{aicity2026track5}
long-horizon traffic future-frame prediction. According to the benchmark
annotations, the test set contains 71 videos and 5194 target frames: 44
front-camera videos, 7 overhead/fixed-camera videos, and 20 vehicle/IP-camera
videos. These labels and counts are reported only to describe the evaluation
set; they are not supplied to the router. At inference, Qwen \cite{bai2025qwen25vl} classifies every
observed clip by matching it to one of three fixed visual-regime prototypes;
the selected regime determines the corresponding prediction branch. A
post-hoc audit against the benchmark-provided reference groups found agreement
for all 71 videos.

Ground-truth future frames for the official test set are not publicly
available, so test metrics are obtained from the challenge evaluation server.
All routing rules and hyperparameters are fixed across the test set; no target
frames or per-video parameter tuning are used during inference.

All experiments are run on one NVIDIA RTX A6000 GPU with 48GB memory, and
full-test inference finishes within one day. The framework is training-free: no
generator weights, adapters, or diffusion sampling schedules are updated.
Table~\ref{tab:implementation_details} summarizes the main implementation
settings.

\begin{center}
\small
\label{tab:implementation_details}
\setlength{\tabcolsep}{4pt}
\renewcommand{\arraystretch}{1.05}
\begin{tabular}{@{}p{0.22\linewidth}p{0.72\linewidth}@{}}
\toprule
Component & Setting \\
\midrule
BDD generator
& Distilled LTX-Video 2B; $1216\times704$ resized to $1280\times720$;
30 FPS; seed 17; 8 steps; guidance 1.0; STG 0; BF16 \\

Video conditioning
& History capped at 41 frames; sequence lengths adjusted to $8n+1$ \\

Text conditioning
& Frozen Qwen2.5-VL-7B-Instruct; structured scene summary from observed
frames; deterministic prompt template \\

View routing
& Frozen Qwen2.5-VL-7B-Instruct; first/middle/last observed frames; fixed
three-way output schema; conservative vehicle/IP fallback \\

Geometry refinement
& Up to 8 history frames, stride 2; point stride 2; splat radius 1;
$\alpha=0.75$; focal scale 0.9 \\

Pose and confidence
& Min.\ 20 ORB matches and 12 PnP-RANSAC inliers; smoothing window 7;
photometric scale 32; confidence EMA 0.65 \\

WTS optical flow
& Farneback at maximum width 426 with OpenCV parameters
$(0.5,3,15,3,5,1.2,0)$ \\

WTS overhead/fixed
& Median-background actor propagation; flow threshold 0.35; RGB thresholds
7/22; component area 20 pixels--2.5\%; blur $\sigma=1.0$; decay 0.985;
maximum scale 12.0 \\

WTS vehicle/IP
& Region-normalized propagation; threshold 1.5; temperature 0.75;
blur $\sigma=3.0$; decay 0.96; maximum scale 6.0 \\
\bottomrule
\end{tabular}
\end{center}

\subsection{Main Results}
\label{sec:main_results}

\begin{center}
\small
\captionof{table}{Official AI City Challenge Track~5 full-test results.}
\label{tab:main_leaderboard}
\setlength{\tabcolsep}{3.5pt}
\renewcommand{\arraystretch}{1.05}
\resizebox{\linewidth}{!}{
\begin{tabular}{@{}clccccccc@{}}
\toprule
Rank & Team & Final & PSNR & SSIM & LPIPS & CLIP & FID & FVD \\
\midrule
1 & Qyn & 76.4866 & 20.1202 & 0.6503 & 0.2463 & 0.9498 & 22.4089 & 21.7907 \\
2 & SSUPER & 76.0385 & 19.7271 & 0.6300 & 0.2487 & 0.9384 & 21.1641 & 19.4627 \\
3 & Latent Painter & 75.1297 & 19.7213 & 0.6504 & 0.2661 & 0.9452 & 26.5176 & 24.7875 \\
4 & CHTTL\_A30 & 74.0544 & 18.8573 & 0.5970 & 0.2814 & 0.9423 & 23.7849 & 24.1105 \\
5 & VGU\_ai\_lab & \textbf{73.2843} & \textbf{19.7360} &
\textbf{0.6472} & \textbf{0.2942} & \textbf{0.9454} &
\textbf{33.6081} & \textbf{29.3483} \\
\bottomrule
\end{tabular}
}
\end{center}

Table~\ref{tab:main_leaderboard} shows that GeoRoute ranks 5th on the official
full-test leaderboard with a final score of 73.2843. Among the top five teams,
our method achieves the second-highest PSNR and an SSIM close to the best
reported value. These results indicate strong preservation of scene layout,
road structure, and low-level appearance, consistent with the goal of anchoring
reliable static regions to observed history frames.

The remaining gap mainly appears in LPIPS, FID, and FVD, which are more
sensitive to perceptual realism and video-level distribution matching. Dynamic
actors, disoccluded regions, and newly synthesized content are deliberately
preserved from the frozen base generator and remain difficult to correct through
static geometry refinement alone. Further improvements therefore require
stronger dynamic-object priors and task-specific video realism.

\subsection{Ablation Study}
\label{sec:incremental_results}

We incrementally add prompt conditioning, geometry refinement, multi-frame
history, confidence-aware blending, and Qwen-assisted \cite{bai2025qwen25vl} routing to the base
predictor. The table summarizes challenge-server development trends rather
than an independent validation-set ablation.

\begin{center}
\small
\captionof{table}{Cumulative ablation of the proposed components.}
\label{tab:incremental_results}
\setlength{\tabcolsep}{2.5pt}
\renewcommand{\arraystretch}{1.0}
\resizebox{\linewidth}{!}{
\begin{tabular}{@{}lcccccc ccccccc@{}}
\toprule
Configuration & Base & Prompt & Geo. & Hist. & Conf. & Route & Final & PSNR & SSIM &
LPIPS$\downarrow$ & CLIP$\uparrow$ & FID$\downarrow$ & FVD$\downarrow$ \\
\midrule
Base prediction
& \checkmark & & & & & & 66.8 & 18.21 & 0.603 & 0.356 & 0.930 & 55.7 & 37.9 \\
Prompt conditioning
& \checkmark & \checkmark & & & & & 67.2 & 18.29 & 0.608 & 0.350 & 0.934 & 53.6 & 36.8 \\
Static geometry
& \checkmark & \checkmark & \checkmark & & & & 69.2 & 18.59 & 0.622 & 0.331 & 0.940 & 47.8 & 31.3 \\
Multi-frame history
& \checkmark & \checkmark & \checkmark & \checkmark & & & 69.4 & 18.61 & 0.620 & 0.331 & 0.940 & 47.2 & 31.2 \\
Confidence blending
& \checkmark & \checkmark & \checkmark & \checkmark & \checkmark & & 71.6 & 19.10 & 0.636 & 0.310 & 0.943 & 39.8 & 30.6 \\
Full routed system
& \checkmark & \checkmark & \checkmark & \checkmark & \checkmark & \checkmark
& \textbf{73.28} & \textbf{19.74} & \textbf{0.647} &
\textbf{0.294} & \textbf{0.945} & \textbf{33.61} & \textbf{29.35} \\
\bottomrule
\end{tabular}
}
\end{center}

\subsection{Qualitative Analysis}
\label{sec:qualitative_analysis}

\begin{center}
\includegraphics[height=0.46\textheight]
{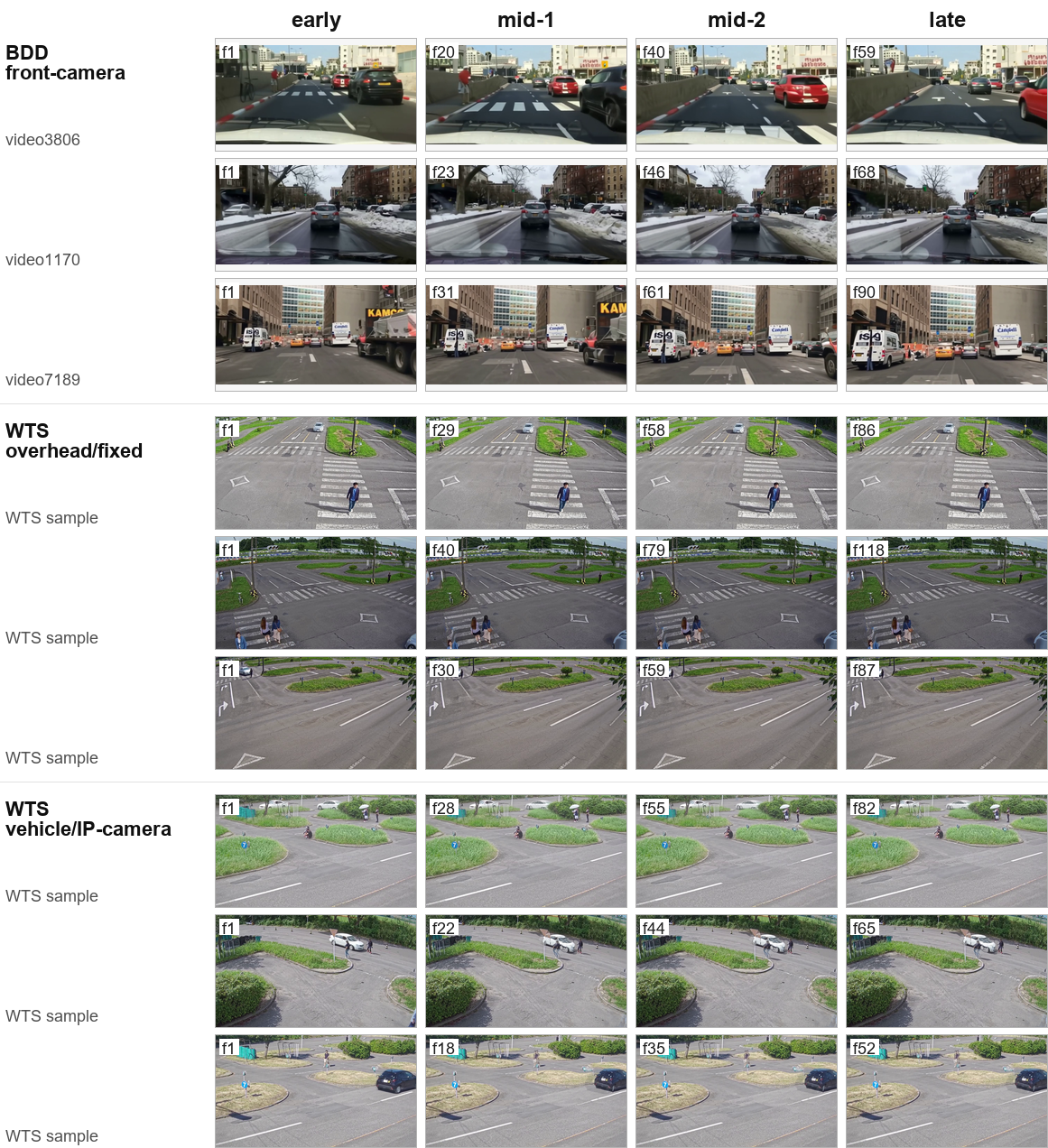}
\vspace{-2mm}
\captionof{figure}{\textbf{Qualitative results across traffic views.}
Predicted future frames from representative BDD front-camera, WTS
overhead/fixed-camera, and WTS vehicle/IP-camera videos. Each row shows
multiple future steps from the same video.}
\label{fig:qualitative}
\vspace{-2mm}
\end{center}

Figure~\ref{fig:qualitative} illustrates the behavior of the complete system
across heterogeneous viewpoints. Front-camera predictions stabilize the global
driving layout, lane boundaries, road geometry, and background structures over
long horizons, while actor motion remains inherited from the base generator.
For WTS overhead/fixed views, the
routed predictor preserves stable backgrounds while propagating localized actor
motion. Vehicle/IP-camera examples also avoid the large global distortions that
can arise from unconstrained flow propagation.

\begin{center}
\includegraphics[width=0.94\linewidth]
{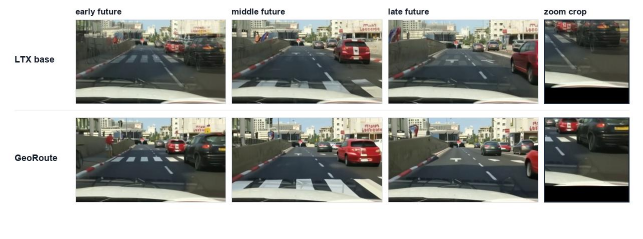}
\vspace{-2mm}
\captionof{figure}{\textbf{Controlled comparison on video3806.}
GeoRoute reduces static-region ghosting near lane markings, road boundaries,
and actor--background boundaries while preserving dynamic regions from the
LTX-Video base prediction \cite{hacohen2024ltxvideo}.}
\label{fig:base_georoute_compare}
\vspace{-2mm}
\end{center}

Figure~\ref{fig:base_georoute_compare} directly compares the base generator with
GeoRoute. The LTX-Video \cite{hacohen2024ltxvideo} sequence blends multiple temporal states around nearby
vehicles and road structures, producing visible afterimages. GeoRoute reduces
the static component of these artifacts by replacing only reliable background
regions with projected history content. Lane markings, curbs, and
actor--background boundaries become cleaner, while dynamic and low-confidence
regions remain close to the generative prediction. This comparison supports
the static-geometry claim but does not imply that GeoRoute re-estimates actor
motion.

\subsection{Failure Analysis}
\label{sec:failure_analysis}

Geometry refinement depends on relative depth, actor masks, and feature-based
alignment. Weak texture, reflections, occlusion, or a poor generated base frame
can therefore reject or misalign the projection. In addition, independently
canonicalized monocular depth maps do not guarantee a common translation scale
for composed multi-history pseudo-transforms. Confidence gating limits the
effect of this approximation but does not make the reconstruction metric.

Recent-priority fusion may also select a newer but less accurate source over an
older well-aligned frame. Dynamic-motion errors remain in the base generator,
and motion propagation may freeze or distort a clip when its observed behavior
does not match the selected regime. Finally, the router assumes that the three
human-defined regimes remain appropriate; automatic discovery of new regimes
is left for future work.

%% file: sec/6_conclusion.tex
\section{Conclusion}

In this work, we presented a geometry-aware, training-free framework for
traffic future-frame prediction. The method combines a text-conditioned video
generator with confidence-aware static refinement and Qwen-assisted routing,
improving static-geometry stability and low-level structural fidelity without
fine-tuning the pretrained model or
changing its sampling schedule. On AI City Challenge Track~5 \cite{aicity2026track5}, our system ranks 5th on the official full-test
leaderboard with a final score of 73.28. These results suggest that explicit
geometric priors and view-aware inference are effective for stabilizing static
structure in training-free traffic video prediction. Dynamic-object motion is
inherited from the base generator rather than explicitly corrected by the
geometry branch. Future work will explore confidence-calibrated learned
routing, stronger dynamic-object motion priors, and tighter integration of
geometry into video generation models.